\documentclass[a4paper,fleqn]{cas-sc}
\usepackage{comment}
\usepackage[numbers]{natbib}
\usepackage[ruled,vlined]{algorithm2e}
\usepackage{subcaption}
\usepackage{graphicx}

\def\tsc#1{\csdef{#1}{\textsc{\lowercase{#1}}\xspace}}
\tsc{WGM}
\tsc{QE}
\tsc{EP}
\tsc{PMS}
\tsc{BEC}
\tsc{DE}

\begin{document}
\def\floatpagepagefraction{1}
\def\textpagefraction{.001}
\shortauthors{Priya Shukla et~al.} 

\title [mode = title]{Robotic Grasp Manipulation Using Evolutionary Computing and Deep Reinforcement Learning 
}




\author[1]{Priya Shukla}[type=editor,
                        auid=000,bioid=1,
                         orcid=https://orcid.org/0000-0002-4163-6238
                        ]
\cormark[1]
\ead{priyashuklalko@gmail.com}

\credit{Conceptualization of this study, Methodology, Software}

\address[1]{Center of Intelligent Robotics, Indian Institute of Information Technology, Allahabad, U.P., INDIA}

\author[2]{Hitesh Kumar}[]
\ead{hkhitesh25@gmail.com }
\address[2]{Indian Institute of Technology (Indian School of Mines), Dhanbad, Jharkhand, INDIA}
\author[1]{G. C. Nandi}[%
  ]
\ead{gcnandi@iiita.ac.in}



\cortext[cor1]{Corresponding author at: Center of Intelligent Robotics, Indian Institute of Information Technology, Allahabad, INDIA}


\begin{abstract}
Intelligent Object manipulation for grasping is a  challenging problem for robots.  Unlike robots, humans almost immediately know how to manipulate objects for grasping due to learning over the years. A grown woman can grasp objects more skilfully than a child because of learning skills developed over years, the absence of which in the present day robotic grasping compels it to perform well below the human object grasping benchmarks. In this paper we have taken up the challenge of developing learning based pose estimation by decomposing the problem into both position and orientation learning. More specifically, for grasp position estimation, we explore three different methods - a Genetic Algorithm (GA) based optimization method to minimize error between calculated image points and predicted end-effector (EE) position, a regression based method (RM) where collected data points of robot EE and image points have been regressed with a linear model, a PseudoInverse (PI) model which has been formulated in the form of a mapping matrix with robot EE position and image points for several observations.
Further for grasp orientation learning, we develop a deep reinforcement learning (DRL) model which we name as Grasp Deep Q-Network (GDQN) and benchmarked our results with Modified VGG16 (MVGG16). Rigorous experimentations show that due to inherent capability of producing very high-quality solutions for optimization problems and search problems, GA  based  predictor performs much better than the other two models for position estimation. For orientation learning results indicate that off policy learning through GDQN outperforms MVGG16, since GDQN architecture is specially made suitable for the reinforcement learning. Based on our proposed architectures and  algorithms, the robot is capable of grasping all  rigid body objects having regular shapes.

\end{abstract}



\begin{keywords}
Grasp Position Mapping \sep Orientation learning \sep Deep Q-Network
\end{keywords}

\maketitle

\section{Introduction}
\subsection{Analysis of previous research}
Grasp Manipulation is a complex multifaced problem which requires substantial learning by humans who after learning can skillfully execute effective grasping in an agile environment. In order for robots to acquire such skill its perception abilities need to be created in order to visually identify the grasp position and orientation for a given object. As it is known that out of main grasping subsystems such as grasp detection system, grasp planning system and grasp control system, the first one is the key point and hence, we have focused on this issue for the present research. 

Traditionally grasp detection and manipulation requires human expert knowledge normally used to be imparted to robots through some analytical models \cite{D1},\cite{c1}. Such analytical models emphasis more on entire grasp system planning. Such approach has severe limitations such as uncertainty issues, like slight pose or position changes of the object require robot to re-plan the entire grasp manipulation task thus causes frequent failures.
On the other hand, learning based model try to make the grasp manipulation problem  analytical model free as far as possible. Several Machine Learning based models and more recently Deep Learning based models are being developed for grasp pose learning \cite{George,Peters}. Sometimes, it gives a better grasp pose but for known object models. This limitation has been tried to be solved in \cite{p27,22,p23} using deep learning techniques incorporating with grasp rectangle concept. They tries to find best grasp rectangle from candidate rectangles, but it is limited to static environment only, as grasp execution with optimal  rectangle is kept in the open-loop as such that there is no feedback for success or failure. In this case, the researcher focuses on visual feedback-based system. This makes the solution adaptable in a dynamic environment. This strategy is called visual servoing. A survey is given in \cite{6} which details the application of visual servoing in  different applications. Recently, visual servoing feedback is incorporated with Convolutional neural network (CNN) \cite{D2,1}. 

In vision based grasping, reachable grasp position plays a vital role in the pick and place task. Image based grasp pose estimation has been solved using Grasp rectangle detection in \cite{22},\cite{p23} which provides the position and orientation in the image coordinate frame. However, this is not sufficient for smooth grasping in the robot configuration space since orientation mapping from image coordinate frame to robot configuration space is not straightforward and a challenging research area. It gives an ambiguous orientation for the same physical orientation. In \cite{O2}, the authors proposed a new representation for orientation as a function of image features, but the axis of rotation was not distinguishable due to its rotation or size decreases. Although its effect is very minimum but impacts on performance. Image based 3D orientation learning has been given by \cite{O1} but it was limited to some few industrial parts. They create data using the Computer Aided Design (CAD) model due to insufficient real images for pose views which itself has limitations. 

It is, therefore, suggested that the orientation learning, for which getting a perfect model is extremely difficult, should be tried with  reinforcement learning techniques. Intuitively, the idea is formulated on the basis of the following observations:
From our childhood to adulthood we are developing grasping skill based on learning from failures. A child initially may fail to grasp an object and she learns the grasping the same object skillfully after several attempts. Grasping is learned over the years for the object to object based on shape and texture. We believe for human object grasping is neither supervised or unsupervised, it is based on experience enriched learning, i.e reinforcement learning. This belief motivated us to formulate a similar learning strategy for robots so that it can grasp objects in a humanoid way. Some related researches are available such as Qt-opt \cite{Qt}, which has addressed the grasping problem utilising huge computational and infrastructural resources. A large number of robots have been deployed with real and synthetic data for training. Here we try to solve the grasping problem with limited available resources and data set and still the performance was quite satisfactory. We do this by decoupling the problem space of grasping in to two-positioning the EE correctly and orientating the same for successful grasping. For position estimate we  develop few methods- mapping from image coordinate frame to robot configuration space using linear regression (LR) method, analytical method PI and GA based optimization method with several fitness functions. The reason for using GA as grasp position determination tool is due to its inherent capability of generating high quality solution. Subsequently, we propose orientation estimation using reinforcement training with our designed model GDQN and benchmark their performance with MVGG16. After extensive experimentation, we observe that the GA based position estimate is the best for EE positioning and GDQN based orientation learning model gives much better performance than existing MVGG16. 

\subsection{Motivation}
\begin{figure}
	\centering
		\includegraphics[scale=.50]{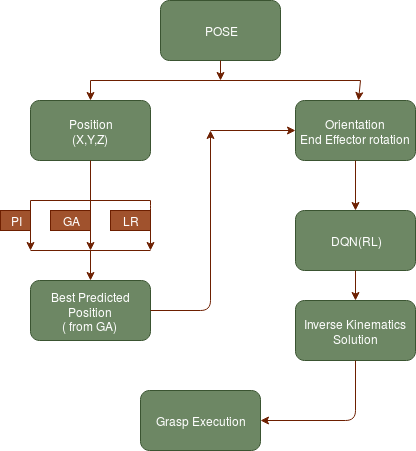}
	    \caption{Grasp Pose estimation and Manipulation Details.}
	\label{fig:flow}
\end{figure}

Grasping requires substantial learning by humans who after learning can skillfully execute effective grasping in an agile environment. In order for robots to acquire such skill its perception abilities need to be created in order to visually identify the grasp position and orientation for a given object. As it is known that out of main grasping subsystems such as grasp detection system, grasp planning system and grasp control system, the first one is the key point and hence, we have focused on this issue for the present research. It involves both position estimation and orientation estimation of the gripper with respect to the given object to be grasped. To solve this problem we emphasis on developing two separate and quite generic models for grasp manipulation. One for grasp position learning and the other for grasp orientation learning, the detail workflow of the entire grasp pose estimation has been illustrated in Fig. \ref{fig:flow}. It illustrates that pose is divided into position and orientation. Position estimation has been solved using three approaches-LR, PI and GA. It has been observed that GA based position estimate, due its inherent capability of producing high quality solutions when configured correctly, is more accurate and hence is more successful in grasping. During experiment we have tried a total of eight different functions, to study which combination gives the best configuration, and analysed their performance in correctly positioning EE over the object to be grasped and the best fitness function has been proposed. Further for orientation learning the current architecture of  DQN \cite{DQN}, \cite{DQN1} has been carefully configured for GDQN and impressive grasp orientation learning results have been obtained which are being illustrated in the subsequent sections. 

One of the  major contributions of the paper is to develop a GA \cite{G3} based gripper position estimation from given detected object by vision camera using two object detection algorithms such as YOLO \cite{OD1} and Faster-RCNN \cite{OD2}. Experiencing that the performance of YOLO (v3) algorithm in a real time environment is better than the Faster-RCNN, the final input to the GA based learning module has been used from YOLO. Objective function for GA optimization has been carefully formulated from calculated value of bounding box center from YOLO algorithm output and corresponding ANUKUL's EE position. The second major contribution is to develop a generic model free approach of gripper orientation estimation using reinforcement learning. 

Philosophically speaking, we extended the self-supervised off-policy based reinforcement learning towards finding correct grasp orientation. Our entire problem encompasses on getting correct bounding box from an object detection algorithm, calculating centre of the box and estimating gripper position using GA and subsequently learning grasp orientation for EE using off-policy based reinforcement learning.

Rest of this paper has been arranged as follows: Section 2 provides problem description and proposed a solution after identifying the limitations of the previous approaches. Section 3 depicts the grasp pose preliminaries and some analytical details about vision, position and orientation learning, state and action representation with reward calculation approaches. Section 4 describes detailed methodologies of each model together with position training, GA formulation for the position estimate algorithm together with orientation learning algorithm. Section 5 illustrates the hardware experimental set up and details about the experiments performed and analyses of results. Section 6 provides conclusion and some insights for future work.

\section{Problem description}
As mentioned in the previous section, image based grasp detection has several limitations. In this section we try to elaborate them in terms  of image based grasp detection and pose estimation. Subsequently, we discuss about our proposed approach to improve the pose estimation (position and orientation) problem by formulating appropriate learning approaches.

\subsection{Limitations of previous approaches}
Image based grasp detection works well in image coordinate space to determine grasp pose but in real world grasp depends on object and EE pose. Image based grasp detection learning plagued with limited number of available datasets \cite{D3},\cite{D4} of some few objects. Dataset creation for pose estimation is quite challenging due to the agile environment. Recently robots itself are being deployed for creating data set for themselves ( Robot for Vision concept).  

\subsection{Proposed Solution}
\begin{figure}
	\centering
		\includegraphics[scale=.50]{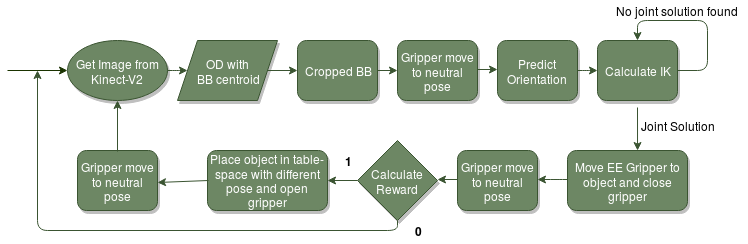}
	    \caption{Grasp Cycle for Orientation Learning.}
	\label{fig:gcycle}
\end{figure}

We try to overcome the limitation of availability of sufficient data set by decomposing the problem space into two - position learning and orientation learning. The position estimate has been done using less data intensive methods like LR, PI and GA where GA produces best position estimation for EE due to its inherent ability to generate quality global solutions.Subsequently, for  grasp orientation learning we allow robot to make trials in some simple steps which are detailed in Fig. \ref{fig:gcycle} where each attempt is considered as one grasp cycle execution and each execution is labelled as success or failure assigning appropriate reward or penalty values respectively.

Each grasp cycle starts by taking image with camera ( here we have used kinect) and applies object detection algorithm on image, once it is detected, bounding box is being cropped and calculates its centroid. Subsequently,hand gripper moves to its neutral pose and predict orientation for centroid of detected object bounding box randomly. If estimated position and predicted orientation is able to find a set of feasible joint solution then it moves its gripper to the target pose and close the gripper otherwise, EE  will predict again its orientation and the iteration continues till it finds feasible solutions in the robot's configuration space. After gripper moves to target pose and close the gripper it moves to the neutral pose.For awarding reward or penalty, the following steps are followed. If object is detected in workspace then it consider as failure and assign rewards as 0 otherwise, gives reward as 1. If calculated reward is 1, then gripper moves to the workspace and keep object on it with different pose and returns to the neutral pose. The whole process is considered as one grasp cycle. It continues during training and learn the correct orientation for pose estimation of gripper. This research is dedicated for correct pose learning by decomposing the problem into both position and orientation learning. In the next section we discuss about basic preliminaries for grasp pose estimation using our proposed methodology.

\section{Grasp pose preliminaries}
In this section, we discuss about vision based pose preliminaries. It requires vision which has been used for learning position as well as orientation. Kinect-V2 camera has been used for vision. Position learning  uses this vision information to predict robot left arm ( we have used one arm out of two) EE position. Subsequently, orientation learning uses both estimated vision and position. This section is further divided into three subsections as described below:

\subsection{Vision}
\begin{figure}
	\centering
		\includegraphics[scale=.50]{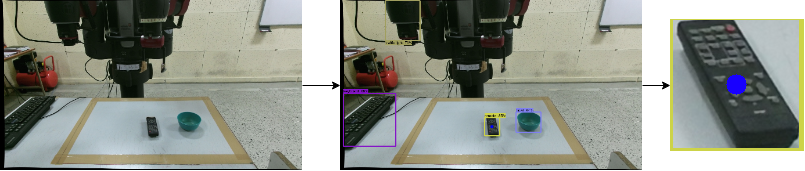}
	    \caption{Vision: Object Detection by YOLO.}
	\label{fig:vision}
\end{figure}

In grasping, vision plays an important role as it provides first hand information towards grasping. Here, object has been identified using an object detection algorithm, which returns bounding box(BB) over the detected object which is shown in Fig. \ref{fig:vision}. Subsequently, bounding box corners are defined in clockwise direction like top-left as A (x\textsubscript1,y\textsubscript1), top-right as B (x\textsubscript2,y\textsubscript1), bottom-right as C (x\textsubscript2,y\textsubscript2) and bottom-left as D (x\textsubscript1,y\textsubscript2), where bounding box width is calculated  as w = x\textsubscript2 - x\textsubscript1 and height calculated as h = y\textsubscript2 - y\textsubscript1. Its center point has been calculated using Eq.(\ref{eq:center}) as:

\begin{equation} \label{eq:center}
 C=(x\textsubscript1+\frac{1}{2}(w),y\textsubscript1+\frac{1}{2}(h)) 
\end{equation}

\subsection{Position learning}
After getting center of the bounding box,the task is to correctly position the gripper. Initially, we create the dataset for 50 points on table top work space designated  for grasping. Here we try to find optimal mapping matrix M (not homogeneous coordinate transformation matrix) using GA, LR and PI. The mapping matrix M maps object vision space with robot's configuration space, and  which is of the shape of $3\times 3$, has been learned using above mentioned methods. Let R be robot left arm EE positions, and I is the center points (for 50 observations) of the detected object bounding box provided by the camera. The mapping relation has been shown in Eq.(\ref{eqn:eq1}). Details of GA, LR and PI techniques for position estimation is illustrated in Methodology section. 

\begin{equation}
R=
\begin{bmatrix}

RX\textsubscript{1} & RX\textsubscript{2} & RX\textsubscript{3} & \hdots & RX\textsubscript{n} \\
RY\textsubscript{1} & RY\textsubscript{2} & RY\textsubscript{3} & \hdots & RY\textsubscript{n}  \\
RZ\textsubscript{1} & RZ\textsubscript{2} & RZ\textsubscript{3} & \hdots & RZ\textsubscript{n} 

\end{bmatrix}_{3\times n}
\end{equation}

\begin{equation}
I=
\begin{bmatrix}
IX\textsubscript{1} & IX\textsubscript{2} & IX\textsubscript{3} & \hdots & IX\textsubscript{n} \\
IY\textsubscript{1} & IY\textsubscript{2} & IY\textsubscript{3} & \hdots & IY\textsubscript{n}  \\
IZ\textsubscript{1} & IZ\textsubscript{2} & IZ\textsubscript{3} & \hdots & IZ\textsubscript{n}  
\end{bmatrix}_{3\times n}
\end{equation}

Let mapping matrix M be defined as :
\begin{equation}
M=
\begin{bmatrix} 
M\textsubscript{(1,1)} & M\textsubscript{(1,2)} & M\textsubscript{(1,3)}\\
M\textsubscript{(2,1)} & M\textsubscript{(2,2)} & M\textsubscript{(2,3)}\\
M\textsubscript{(3,1)} & M\textsubscript{(3,2)} & M\textsubscript{(3,3)}
\end{bmatrix}_{3\times 3}
\end{equation}
\begin{equation}\label{eqn:eq1}
\underset{3\times n}{R}=\underset{3\times 3}{M}\times \underset{3\times n}{I}
\end{equation}
Where:
\begin{itemize}
  \item R ($3\times n$) is Anukul left arm EE positions.
  \item I ($3\times n$) is detected objects centroid.
  \item M ($3\times 3$) is mapping matrix.
  \item n is the no. of observations.
\end{itemize}

\subsection{Orientation learning formulation}
Orientation learning uses the calculated ( as described earlier) position to predict EE orientation. Discrete Reinforcement Learning (DRL) needs state, action and reward. State (S) has been defined as a detected object BB as an image and action space has been defined as based on action mask size. Action mask has been defined as the number of actions. Here we have taken 3, 12 and 24 action spaces, where 3 action space includes 0, 45 or 90 EE orientation, 12 action space includes EE orientation in the difference of 15 degree and 24 action space includes EE orientation in the difference of 7.5 degree. State, action and reward have been formulated below:

\subsubsection{State representation}
Detected object bounding box cropped from an RGB image of size 1920x1080 pixels at time t. RGB image has been taken from an external camera Kinect-V2, which is fixed in front of Anukul robot to view tablespace. Workspace in the table has been defined as 63x86 cm\textsuperscript{2}. For state detected object bounding box cropped from RGB image, cropped image is then converted to 84x84x1 for GDQN and 200x200x3 for MVGG16 network models.

\subsubsection{Action representation}
Orientations have been denoted as nA. It tells the possible action value for orientation in the specified workspace for gripper orientation. Here, $\pi$ has been considered as the workspace for orientation learning. In Anukul robot EE gripper can rotate upto $2\pi- 10$. So rotation difference for the specified action has been calculated as:

\begin{equation}\label{Eq:actionp1}
k = \frac{\pi}{nA}
\end{equation}
\begin{equation}\label{Eq:actionp2}
\Theta = (0/1/2/3/4/...../nA)\times k 
\end{equation}

In this work we represented our proposed approach with 3, 12 and 24 possible action space. For 3 action space EE orientation range is 0 to 90 whereas for 12 and 24 action space EE orientation range is 0 to 180. For 3 action $\pi$ needs to replace with $\frac{\pi}{2}$ and rest is the same.

\subsubsection{Reward calculation}
Here, reward has been calculated using images of  before grasping and after grasping states. In the after grasping image state, object detection has been applied to detect object. If object detection returns object coordinate then reward will be set to 1 otherwise it is set to 0. Further details are shown in the  Algorithmic form \ref{alg:reward}. With discrete values in reward function learning takes longer time but it returns more accurate results.
\\
\begin{algorithm}[H]
\SetAlgoLined
\KwResult{Return Reward}
    initialize:coordinates[0] == 0 and coordinates[1] == 0;
 
    \eIf{OD(image)}{
         Reward=1\;
    }{
   Reward=0\;
 }
 \caption{Reward function()}
 \label{alg:reward}
\end{algorithm}

Grasping task has been considered as an agent learning from the environment with discrete interactions to obtain best optimal grasp orientation angle for the given object so as to find the most optimal EE orientation for a given object. At each discrete interaction with the environment at time step t = 0, 1, 2, 3, . . . . . n, agent receives some form of environmental state as its current state s\textsubscript{t} $\in$ S, which in our case is the image patch of the target object in the environment, and performs an action
$\theta$\textsubscript{t} $\in$ A, which in our case is the EE orientation for the corresponding target object and as a consequence of this action, it receives a reward r\textsubscript t  (s\textsubscript t, $\theta$\textsubscript{t}) $\in$ R and observe the new environmental state s\textsubscript{t+1} as its next state. Although the set of states S  could be very large, still it has the finite number of elements and sets of actions and rewards are also formulated to be finite. Subsequently, we formulated agents interacting with the environment as a finite Markov Decision Process(MDP) in which agent and environment interact for $\tau$ time steps per episode giving rise to the following sequence as history H.
H = (s\textsubscript {0} , $\theta$\textsubscript{0} , r\textsubscript{0} , s\textsubscript{1}, $\theta$\textsubscript{ 1}, r\textsubscript{1}, s\textsubscript{2}, $\theta$\textsubscript{2}, r\textsubscript{2}, s\textsubscript{ 3}, . . .  . . . . . . . . ,r\textsubscript{$\tau$-2},  s\textsubscript{$\tau$-1}, $\theta$\textsubscript{$\tau$-1}, r\textsubscript{$\tau$-1}, s\textsubscript{$\tau$}, $\theta$\textsubscript {$\tau$}, r\textsubscript{$\tau$}).
We make the standard assumption that future rewards are discounted by a factor called $\gamma$ time step, so that the discounted future rewards at time step t will be :

\begin{equation} \label{eqreward}
R\textsubscript{t} =\sum_{t\textsuperscript{`}=t}^{\tau}\gamma^ {t\textsuperscript{`}-t} r\textsubscript{t}(s\textsubscript{ t},\theta\textsubscript{t})
\end{equation}

Now the goal of the agent is to find an optimal policy $\pi$\textsubscript{$\Theta$}  parametrized by $\theta$  which maximizes the expected future reward i.e.

\begin{equation}  \label{eqn:5}
\pi\textsubscript{$\Theta$}(\theta\mid s ) = \max_{\pi} E  \sum_{t\textsuperscript{`}=t}^{\tau} \gamma ^{t\textsuperscript{`}-t} r\textsubscript{t} (s\textsubscript{t},\theta\textsubscript{t} )
\end{equation}

\begin{equation} \label{eqn:6}
\pi\textsubscript{$\Theta$}(\theta\mid s ) = \max_ {}Q\textsubscript {$\Theta$}(s,\theta)
\end{equation}

Where Q\textsubscript{$\Theta$}(s,$\theta$) is the additive discounted future rewards by following a policy $\pi$ and parameterised by $\Theta$. To find an optimal policy, two models have been used to learn EE orientation which are GDQN and MVGG16. 
Model details have been explained in the Methodology section and their impacts on learning have been described in the Robotic Experimental Results section.

\section{Methodology}

\subsection{Position training techniques}
Position mapping has been solved by obtaining optimal mapping matrix M using an evolutionary algorithm (GA), Regression models (LR) and statistical tool (PI). By using these mapping techniques robot arm is able to reach the object at the graspable position. During data collection, we have used left arm EE (the other arm could also be used without loss of generality). GA has been solved as an optimization problem whereas LR has been solved with regression models for each axis (X, Y and Z) and PI is solved by analytical/statistical tool. All methods are capable of reducing the error, with various degrees, between actual and predicted EE position. Here, we have used quadratic loss function for position estimation ,which has been calculated as shown in Eq.(\ref{eqn:eqloss}). Sufficient Details of each method is given below:

\begin{equation}\label{eqn:eqloss}
    Loss (R\textsubscript{p})= \sum_{i=1}^{n}({R\textsubscript {actual\_position} }- {R\textsubscript {predicted\_position}})^ 2
\end{equation}

\subsubsection{Genetic algorithm}
\begin{figure}
	\centering
		\includegraphics[scale=.50]{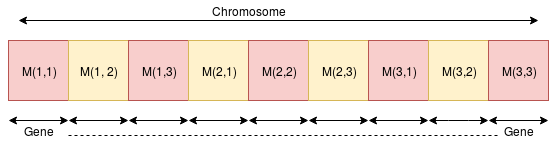}
	    \caption{Chromosome}
	\label{fig:gene}
\end{figure}

Genetic algorithm, as evolutionary computation technique, has been used as an optimizer for determining  mapping matrix (M). The expression for M with image matrix and robot EE position matrix for n observations has been shown in Eq.\ref{eqn:eq1} and the procedure of determining position has been illustrated in the Algo.\ref{alg:Po_Ga}. Here we have proposed eight different fitness functions for mapping matrix M between I and R, where error function is considered as fitness function. Subsequently, computed their accuracy in testing time with Root Mean Square Error (RMSE). For GA, the mapping matrix M which is being size of $3\times 3$ is converted into $1\times 9$ size of matrix. It represents chromosome of length 9 where each element is considered as gene which is detailed in Fig. \ref{fig:gene}. 

\begin{table}
    \centering
    \caption{Fitness Functions}
    \label{tab:Table1}
    \begin{center}
    \begin{tabular}{|c|c|c|}
    \hline
    \textbf{S.No.} & \textbf{(Function)} & \textbf{Error(RMSE) } \\
    \hline
    1 &$E=\sum_{i=1}^{n}R-MI$&0.00132123\\\hline
    2 &$E=\sum_{i=1}^{n}MI-R$&0.00134930\\\hline
    3 &$E=\sum_{i=1}^{n}RI\textsuperscript{-1}-M$&0.00000207\\\hline
    4 &$E=\sum_{i=1}^{n}M-RI\textsuperscript{-1}$&0.00000862\\\hline
    5 &$E=\sum_{i=1}^{n}R\textsuperscript{-1}M-I\textsuperscript{-1}$&0.01164770\\\hline
    6 &$E=\sum_{i=1}^{n}I\textsuperscript{-1}-R\textsuperscript{-1}M$&0.01177603\\\hline
    7 &$E=\sum_{i=1}^{n}M\textsuperscript{-1}R-I\textsuperscript{-1}$&0.02713571\\\hline
    8 &$E=\sum_{i=1}^{n}I\textsuperscript{-1}-M\textsuperscript{-1}R$&0.02713083\\\hline
    \end{tabular}
    \end{center}
\end{table}

Detailed analysis of fitness performance has been tabulated in Table \ref{tab:Table1}. Fitness function 1 and 2 is free from matrix non-invertability issue, related to the inverse of the non square matrix, so their computational complexity is reduced, however, with relatively poor performance. For fitness function 3 and 4 computational complexity is increased but they perform much better than 1 and 2. In fact fitness function 3, $E=\sum_{i=1}^{n}RI\textsuperscript{-1}-M$ outperforms all eight fitness functions we try and provides the best error convergence. 
 We, therefore, use Eq.(\ref{eq8}) as the selected fitness function of our choice for further training and testing purposes. 

\begin{equation}\label{eq8}
E=\sum_{i=1}^{n}RI\textsuperscript{-1}-M
\end{equation}

Mapping matrix M is obtained by using Algo.\ref{alg:Po_Ga}, where n is the number of observations it has taken to minimize graspable position prediction error for robot arm EE. Position learning using GA algorithm has been detailed in   Algo.\ref{alg:Po_Ga}. In the previous work \cite{p21} fitness function has been used as $E=\sum_{i=1}^{n}R-MI$ but in this work fitness function has been improved as it is defined in Eq.(\ref{eq8}). Some parameters which are used illustrated below: 
\begin{itemize}
    \item N\textsubscript{w} $\leftarrow$ no. of weights.
    \item S\textsubscript{p} $\leftarrow$  solution per population size.
    \item N\textsubscript{p} $\leftarrow$  population size.
    \item P $\leftarrow$  Parents.
    \item Off\_X $\leftarrow$  offspring crossover.
    \item Off\_M $\leftarrow$  offspring mutation.
    \item N\textsubscript{offsp} $\leftarrow$  offspring size.
\end{itemize}

Here, position prediction error has been estimated as detailed in Eq.(\ref{eqn:eqloss}), where n is the test observations in robot configuration space from image coordinate space.

\begin{algorithm}
    \SetAlgoLined
    \SetKwInOut{Input}{Input}
    \SetKwInOut{Return}{Return}
    \Input{$R_{3 \times n}$, $I_{3 \times n}$}
    \Return{$T_{3 \times 3}$}
    $Initialize: N_{w}, S_{p} , G, M, Fit, Best\_Fit, \mathit{New\_pop} $\;
    $N_{p} = N_{w} \times S_{p}$\;
    \For{i $\leftarrow$ 1 $\KwTo$ G}
    { Fit,$Best\_Fit$ $=$ $GA\_cal\_pop\_fitness$($R$,$I$,$\mathit{New\_pop}$)\; 
        P $=$ $GA.select\_matting\_pool(\mathit{New\_pop}$, Fit, M)\;
        $\mathit{Off\_X}$ $=$ $GA.crossover$(P,$N\textsubscript{offsp}$,$N\textsubscript{w})$\;
        $\mathit{Off\_M}$ $=$ $GA.mutation(Off\_X)$\;
        $New\_pop[0:P.shape[0], :]$ $=$ P\;
        $New\_pop[P.shape[0]:, :] $ $=$ $Off\_M$\;} 
\textbf{Return}~ $New\_pop [best\_fitness\_value]$ \;
\caption{Position\_GA}
\label{alg:Po_Ga}
\end{algorithm}

\subsubsection{Linear Regression}
Here, we use three regression models separately along x ,y and z axes with  n observation data. Regression model  for x-axis is detailed as: X\textsubscript{1} -image points along x-axis and Y\textsubscript{1}- robot positions in x-axis. Let $\omega\textsubscript{x\textsubscript{1}}$ is defined as slope or gradient and $\omega\textsubscript{x\textsubscript{0}}$  is defined as intercept. Here, we use 
\begin{equation}
    Y=\omega\textsubscript{x\textsubscript{0}}+\omega\textsubscript{x\textsubscript{1}}X
\end{equation}
as regression line and the normal equation is given as:
\begin{equation}
    \omega\textsubscript{x\textsubscript{1}} = \frac{n\sum X\textsubscript{1}Y\textsubscript{1}  - \sum X\textsubscript{1}\sum Y\textsubscript{1}}{
    n \sum X\textsubscript{1}^2 - (\sum X\textsubscript{1})^2} 
\end{equation}

\begin{equation}
    \omega\textsubscript{x\textsubscript{0}} = \frac{\sum X\textsubscript{1}^2  \sum Y\textsubscript{1}  - (\sum X\textsubscript{1})\sum X\textsubscript{1} Y\textsubscript{1}}{
    n \sum X\textsubscript{1}^2 - (\sum X\textsubscript{1})^2} 
\end{equation}

Similarly, regression model for y-axis is detailed as: X\textsubscript{2}- image points in y-axis and Y\textsubscript{2}- robot positions in y-axis. Let $\omega\textsubscript{y\textsubscript{1}}$ is defined as slope or gradient and $\omega\textsubscript{y\textsubscript{0}}$ is defined as intercept.

\begin{equation}
    \omega\textsubscript{y\textsubscript{1}} = \frac{n\sum X\textsubscript{2}Y\textsubscript{2}  - \sum X\textsubscript{2}\sum Y\textsubscript{2}}{
    n \sum X\textsubscript{2}^2 - (\sum X\textsubscript2)^2} 
\end{equation}

\begin{equation}
    \omega\textsubscript{y\textsubscript{0}} = \frac{\sum X\textsubscript{2}^2  \sum Y\textsubscript{2}  - (\sum X\textsubscript{2})\sum X\textsubscript{2} Y\textsubscript{2}}{
    n \sum X\textsubscript{2}^2 - (\sum X\textsubscript{2})^2} 
\end{equation}

with  \begin{equation}
    Y=\omega\textsubscript{y\textsubscript{0}}+\omega\textsubscript{y\textsubscript{1}}X
\end{equation}

and regression model for z-axis is detailed as: X\textsubscript{3}- image points in z-axis and Y\textsubscript{3}- robot positions in z-axis. Let $\omega\textsubscript{z\textsubscript{1}}$ is defined as slope or gradient and $\omega\textsubscript{z\textsubscript{0}}$ is defined as intercept.

\begin{equation}
    \omega\textsubscript{z\textsubscript{1}} = \frac{n\sum X\textsubscript{3}Y\textsubscript{3}  - \sum X\textsubscript{3}\sum Y\textsubscript{3}}{
    n \sum X\textsubscript{3}^2 - (\sum X\textsubscript{3})^2} 
\end{equation}

\begin{equation}
    \omega\textsubscript{z\textsubscript{0}} = \frac{\sum X\textsubscript{3}^2  \sum Y\textsubscript{3}  - (\sum X\textsubscript{3})\sum X\textsubscript{3} Y\textsubscript{3}}{
    n \sum X\textsubscript{3}^2 - (\sum X\textsubscript{3})^2} 
\end{equation}

with  \begin{equation}
    Y=\omega\textsubscript{z\textsubscript{0}}+\omega\textsubscript{z\textsubscript{1}}X
\end{equation}

All above illustrated models have been used to predict robot EE position in robot configuration space using image coordinate positions in Cartesian space with Quadratic loss function as calculated using Eq. (\ref{eqloss1}).

\begin{equation}\label{eqloss1}
    Loss (Y\textsubscript{i})= \sum_{i=1}^{3}({Y\textsubscript{actual\_position\_i} }- {Y\textsubscript {predicted\_position\_i}})^ 2
\end{equation}

\subsubsection{PseudoInverse}
From Eq.(\ref{eqn:eq1}), we can infer that the  Robot configuration space mapping matrix M can be obtained easily if robot matrix R and image matrix I are square matrices and I is nonsingular. But for the grasping environment matrices size would depend on number of observations taken and it is highly unlikely that such constrains would be obeyed always for n number of observations. Hence, whenever, M is computable, we compute it using the set of Eq.(23)-(25) as illustrated below:

\begin{equation*}
  \underset{3\times n}{\mathrm{R}}=\underset{3\times 3}{\mathrm{M}} \times \underset{3\times n}{\mathrm{I}}
\end{equation*}
\begin{equation}
  \underset{3\times n}{\mathrm{R}}\times \underset{n\times 3}{\mathrm{I}}\textsuperscript{T} =\underset{3\times 3}{\mathrm{M}} \times \underset{3\times n}{\mathrm{I}} \times \underset{n\times 3}{\mathrm{I}}\textsuperscript{T}
\end{equation}
\begin{equation}
  \underset{3\times n}{\mathrm{R}}\times \underset{n\times 3}{\mathrm{I}}\textsuperscript{T} \times (\underset{3\times n}{\mathrm{I}} \times \underset{n\times 3}{\mathrm{I}}\textsuperscript{T})\textsuperscript{-1} =\underset{3\times 3}{\mathrm{M}}
\end{equation}

Now
\begin{equation} \label{eq3}
  \underset{3\times 3}{\mathrm{M}} = \underset{3\times n}{\mathrm{R}}\times \underset{n\times 3}{\mathrm{I}}\textsuperscript{T} \times (\underset{3\times n}{\mathrm{I}} \times \underset{n\times 3}{\mathrm{I}}\textsuperscript{T})\textsuperscript{-1}
\end{equation}

\subsection{Orientation training details  }
Some important aspects of orientation learning are discussed in this section with orientation learning algorithm, GDQN and modified VGG network. During training for each grasp attempt a complete grasp cycle is considered which is illustrated in Fig. \ref{fig:gcycle}.

\subsubsection{Orientation learning algorithm}
Here, we have illustrated learning orientation using grasp cycle. Details of each step have been discussed in Algo.\ref{alg:Orient_RL}, which includes the state(s) as the cropped bounding box of the detected object image from an input image by Kinect-V2 camera and action($\Theta$) as defined from Eq. 5 and 6. As it is written in \cite{BK}, verified EE can rotate up to 350 degrees.  For three action rotation range is considered as $ \frac{\pi}{2}$. Subsequently for 12 and 24 action rotation range is considered as $\pi$. We have used replay memory of 1000 to store each grasp attempt as initial data and after learning it gets updated further. Action selection depends on epsilon greedy action policy function. It considers $\epsilon$ decay with the current episode (CE) and current state (CS). Initially, action selection is random but after convergence of $\epsilon$, action selection has been predicted by the learned model.

\begin{algorithm}
Initialize: $RM$, TOC, TMC, nEpisodes, $\epsilon\textsubscript{f}$ \;
$\epsilon\textsubscript{d} = (1.0-\epsilon\textsubscript{f}/nEpisodes)$ \; 
$S \leftarrow I\textsubscript{$x \times y$}, nA $\;
$env = Baxter\_Env() $\;
$model ,model\textsubscript{t}=grasping\_model()$\; 

\eIf{!Model}{ Load=True;}{Load=False;} \
 \SetKwFunction{FMain}{Train}
    \SetKwProg{Fn}{Function}{:}{}
    \Fn{\FMain{}}
    {
        env.reset = s , s = preprocess(s) \;
        \While{$CE < nEpisodes$}
        {
            $a=epsilon\_greedy\_action\_policy(s,CE)$\;
            $k = \frac{\pi}{nA}$\;
            $o=a \times k$ \;
            $\epsilon = \epsilon - \epsilon\textsubscript{d}$ \;
            $s\textsubscript{t+1},r = env.step(o)$ \;
            $RM.append(s,o,r,s\textsubscript{t+1},done)$ \;
            \If{$CE\geq TOC$}{deepQlearn();}\
                \If{(CE \% TMC  == 0)}
                    {$ COPY(\theta ,model\textsubscript{t}) $ \;
                } 
            $s=s\textsubscript{t+1}$ \;
            $score=score+r$ \;
            $CE=CE+1$ \;

        }
   }
   
\textbf{End Function}
\caption{Orientation\_Learning} 
\label{alg:Orient_RL}
\end{algorithm}

\subsubsection{Models Architecture}

\subsubsection*{Our custom Model GDQN }
\begin{figure}
	\centering
		\includegraphics[scale=.50]{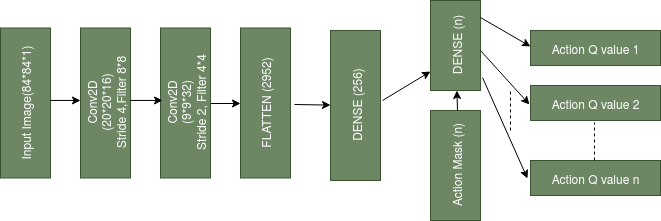}
	    \caption{Grasp Deep Q Network.}
	\label{fig:gdqn}
\end{figure}

The very first model which is inspired by \cite{atari} and model that we used to approximate the Q-Function is a densely connected deep neural network with a total of 673k trainable parameters. The architecture of the network is shown in Fig.\ref{fig:gdqn}  with layer specification using stride and filter. It takes a grayscale image of size 84$\times$ 84 as input and process it through two convolutional layers followed by two fully connected layers and outputs a 3- dimensional vector for action. The first convolutional layer convolves 16, $8\times 8$ filters with stride 4 and uses ReLU as the activation function. The second convolutional layer convolves 32, $4\times 4$ filters with stride 2 and also uses ReLU as the activation function. In this experiment, we used RMSProp as the optimiser with lr=0.00025 with minibatch of size 8.

\subsubsection*{Modified VGG Model}
\begin{figure}
  \centering
		\includegraphics[scale=.50]{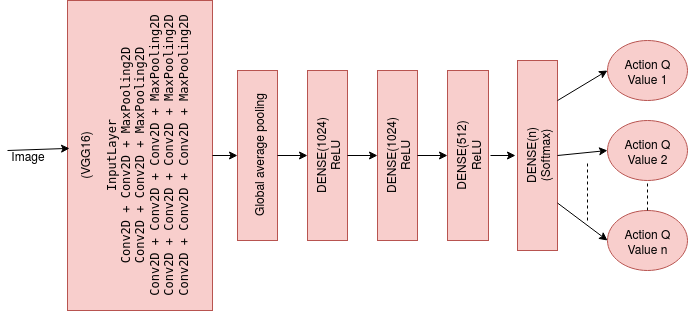}
	    \caption{Modified MVGG16 network.}
	\label{fig:mvgg}
\end{figure}

Here we have used a pre-trained vgg16 model trained on COCO dataset \cite{COCO} with modified fully connected layers. The network architecture is shown in Fig.\ref{fig:mvgg} with used activation function in each layer. We froze the first twenty layers and thereafter add four Fully Connected layers with different number of neurons like 1024, 512, and 3 respectively and activation function ReLU for first three FC layers and softmax for the last one. The total number of trainable parameters thus become 2M. In this experiment we again used RMSProp as the optimiser with lr=0.00025 with minibatch of size 8.

\subsubsection{Loss Function}
During Orientation training, we have used Huber loss. It negotiates between absolute and squared loss function \cite{Huber}.

      
 $
 L_{\delta}(Q\textsubscript {$\Theta$}(s\textsubscript{i},\theta\textsubscript{i}), y\textsubscript{i}) = 
 \begin{cases} 
 \frac{1}{2} * (K )^2, \quad if\hspace{1.5pt} |K| \leq \delta, \\ 
 \delta * (|K| -  \frac{1}{2} * \delta),   \quad otherwise 
 \end{cases} 
$

where:
\begin{itemize}
    \item $K=Q^{w\textsubscript{i}}\textsubscript {$\Theta$}(s\textsubscript{i},\theta\textsubscript{i}) -y\textsubscript{i}^{\overline w\textsubscript{i}}$ 
    \item $w\textsubscript{i}$ $\leftarrow$ network parameters.
    \item $\overline w\textsubscript{i}$ $\leftarrow$ target network parameters.
\end{itemize}

\section{Experimental Results with Robot in the loop}
All the experiments have been implemented on real time robot Anukul. Pose estimation results have been divided into two major parts-position learning and orientation learning. Above mentioned grasp cycle execution has been shown in Fig. \ref{fig:OLsteps} and details related to Robotic Experimental Setup have been illustrated below:

\begin{figure}
	\centering
		\includegraphics[scale=.40]{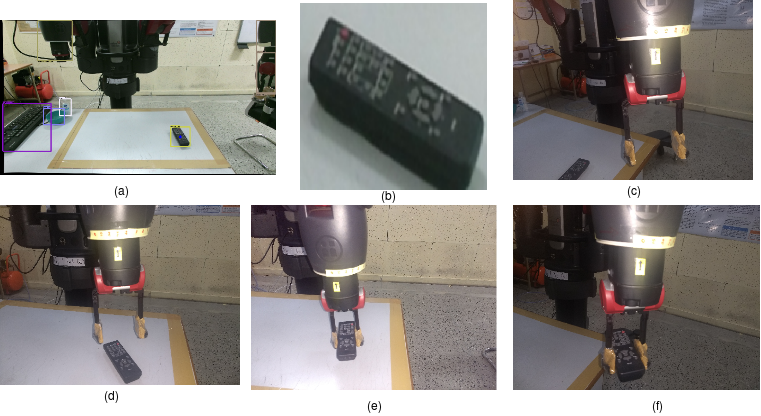}
	    \caption{Grasp execution cycle- It shows basic steps to grasp an object (a) represents detected object with center. (b) is image patch of detected bounding box. (c) is defined as neutral pose. (d) predicts orientation and move toward object. (e) close the gripper. (f) moves object to the neutral pose and calculates reward.}
	\label{fig:OLsteps}
\end{figure}

\subsection{Robotic Experimental Setup}
\begin{figure}
    \centering
        \includegraphics[scale=.50]{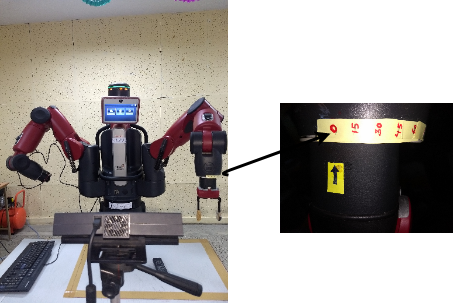}
        \caption{{Robotic Experimental Setup}}
  \label{fig:setup}
\end{figure}

In this research for proposed approach verification, Anukul Research Robot has been used for grasping as a manipulator task and Kinect-V2 has been used for vision. Robotic Experimental Setup has been shown in Fig. \ref{fig:setup}. Anukul is a research robot created by Rethink Robotics, and is different from the traditional industrial robot as it is called as COBOT (COllaborative roBOT), which can work together with humans as a team in the warehouse for any kind of manipulation work based on human interaction. Its arm movement is elastic so it safe to work with it for humans. It has two arms, each is configured with seven degrees of freedom. There are three internal cameras - two on both the arm EE and one on top of the head. Camera maximum resolution is fixed at 1280 $\times$ 800 pixels with 30 Frame Per Second (FPS) and focal length is 1.2 mm. Obtained image with these configurations are not enough for object detection. So  we used an external camera Kinect-V2 external camera with the following configurations:
640 $\times$ 480 pixels @ 30 Hz for RGB camera whereas 640 $\times$ 480 pixels @ 30 Hz for IR depth-finding camera \cite{Kinect}. This robot model is equipped with both electrical and vacuum grippers. Throughout all the experimentation we have used the left arm equipped with electrical gripper (the other arm could also be used without loss of generality). For the Setup Kinect -V2 is placed in front of Anukul robot so that it can see the table workspace where boundary has been shown by brown colour. Orientation verification has been done by marking angles on top of the left arm EE. During experiments, we have used regular shape objects because grasping using bounding box centroid is well suited for such objects.

\subsection{Position Learning}
 Position learning has been tested using three different approaches as discussed in the methodology section. Results for obtaining optimal mapping matrix with eight different fitness function (error function) has been presented  in the Table \ref{tab:Table1}. $E=\sum_{i=1}^{n}RI\textsuperscript{-1}-M$ fitness function outperforms all other fitness functions. Along with GA, position estimation is also done using regression and PI. But in all cases, GA outperforms both regression and PI. This is perhaps due to the inherent ability of GA to produce quality solutions( ref Schema theory). The comparison is shown in Table \ref{tab:Table2}.

\begin{figure}
\minipage{0.32\textwidth}
  \includegraphics[width=\linewidth]{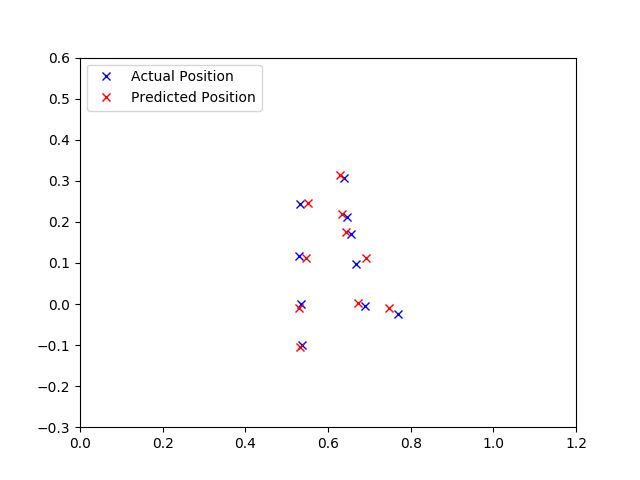}
  \caption{GA Predicted/Actual Position}\label{fig:GA_test}
\endminipage\hfill
\minipage{0.32\textwidth}
  \includegraphics[width=\linewidth]{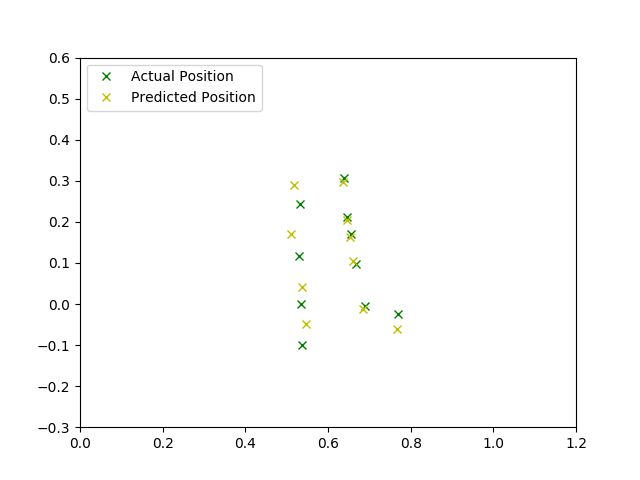}
  \caption{LR Predicted/Actual Position}\label{fig:LR_test}
\endminipage\hfill
\minipage{0.32\textwidth}%
  \includegraphics[width=\linewidth]{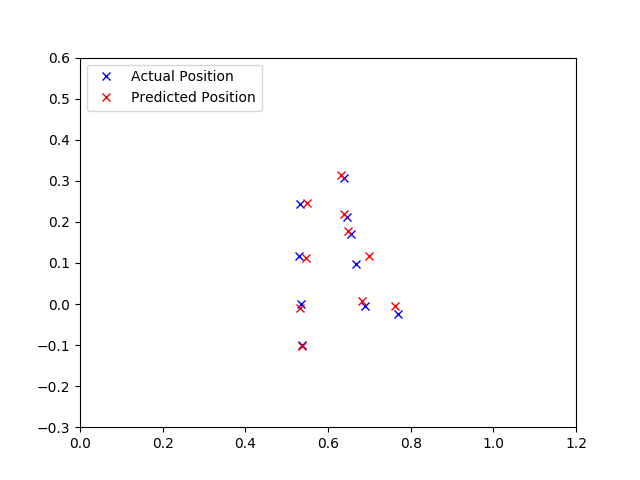}
  \caption{PI Predicted/Actual Position}\label{fig:PI_test}
\endminipage
\end{figure}

\begin{table}
\caption{Prediction Error Comparison}
\label{tab:Table2}
\begin{center}
\begin{tabular}{|c|c|c|}
\hline
\textbf{S.No.} & \textbf{Technique} & \textbf{Error(RMSE) } \\
\hline
1 &Genetic algorithm&0.0095819\\\hline
2 &Linear regression&0.0201485\\\hline
3 &PseudoInverse&0.0106281\\\hline
\end{tabular}
\end{center}
\end{table}

\subsubsection{Genetic Algorithm}
 For GA among all tried solutions as mention in Table \ref{tab:Table1}, $E=\sum_{i=1}^{n}RI\textsuperscript{-1}-M$ fitness function used for final position mapping as it works well in  converging errors at 0.00000207 for 40 observations which is shown in Fig. \ref{fig:Fit}  and Root Mean Square Error (RMSE) for 10 observation is 0.00958199. Position Mapping by GA is shown in Fig. \ref{fig:GA_test} with a blue coloured and red coloured symbol, where the red colour symbol represents predicted EE position and blue colour actual EE position.  Overlapping of cross symbols in Fig. \ref{fig:GA_test} infers error in EE position prediction which is considered negligible.

\subsubsection{Linear regression}
Regression has solved the position mapping problem by obtaining fit value in each direction as X, Y and Z axes respectively. Here, we have taken 50 observations and shuffled the data. 40 observations have been used for training and 10 for testing. Test results are shown in Fig. \ref{fig:LR_test}. Here, we used the yellow coloured and green coloured cross symbol, where the yellow colour symbol represents predicted EE position and green colour actual EE position. Overlapping of cross symbols in Fig. \ref{fig:LR_test} infers error in EE position prediction which is considerably high as compared to GA as RMSE for 10 test observation is 0.0201485.

\subsubsection{PseudoInverse}
Here, pseudoinverse has been verified to get mapping matrix for left arm EE (R) from image points (I). We can get mapping matrix M by $M=RI\textsuperscript{-1}$ but R and I are non-invertible matrices here so inverse in possible for those matrices. So we used Eq.(\ref{eq3}) to get the mapping matrix which is equivalent to inbuilt PI function. 
Position Mapping by PI is shown in Fig. \ref{fig:PI_test} with a blue coloured and red coloured symbol, where the red colour symbol represents predicted EE position and blue colour actual EE position.  Overlapping of cross symbols in Fig. \ref{fig:PI_test} infers error in EE position prediction. It works better than LR but GA outperforms among three approaches as shown in the Table \ref{tab:Table2}. 

\begin{figure}
\minipage{0.48\textwidth}
  \includegraphics[width=\linewidth]{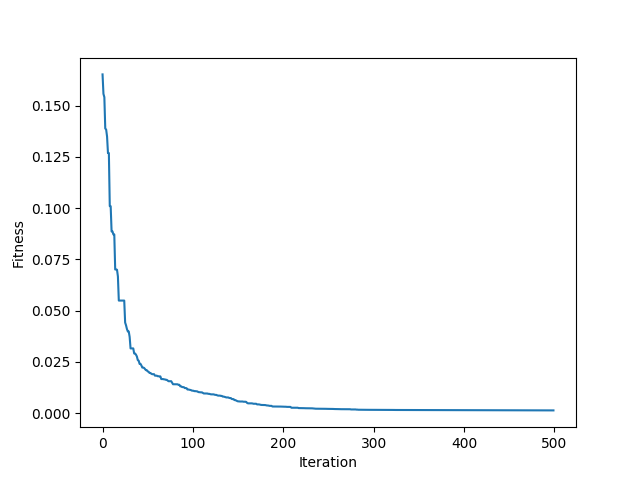}
  \caption{Fitness Function}\label{fig:Fit}
\endminipage\hfill
\minipage{0.48\textwidth}
  \includegraphics[width=\linewidth]{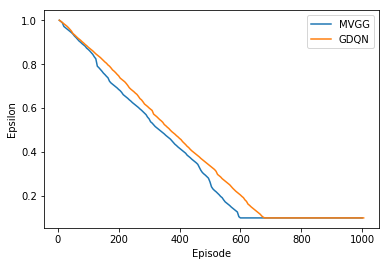}
  \caption{Epsilon decay}\label{fig:epsilon}
\endminipage\hfill
\end{figure}

\subsection{Orientation Learning}
\begin{figure}
\minipage{0.48\textwidth}
  \includegraphics[width=\linewidth]{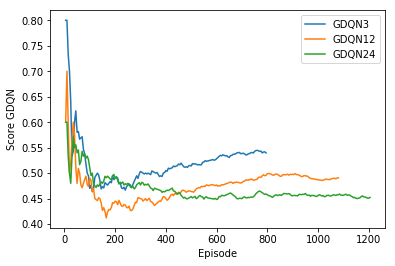}
  \caption{Score Vs Episode for GDQN}\label{fig:Score_GDQN}
\endminipage\hfill
\minipage{0.48\textwidth}
  \includegraphics[width=\linewidth]{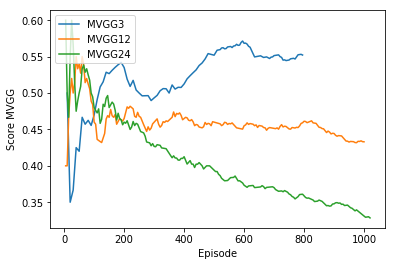}
  \caption{Score Vs Episode for MVGG16}\label{fig:Score_MVGG}
\endminipage\hfill
\end{figure}

For orientation learning three different action spaces have been used for training, these are defined as 3-actions, 12-actions and 24-actions. During grasp orientation learning EE predicts angle for grasping and same prediction approach has been used to place the object in table workspace. For 3-action prediction range is set as 0 to 90 and predicted action has been considered as 0, 45 and 90. The same strategy has been applied to 12-action and 24-action with prediction range from 0 to 180 where action as an EE orientation has been calculated using Eq. (\ref{Eq:actionp1}) and (\ref{Eq:actionp2}).
Here GDQN and MVGG16 models have been trained with all three action spaces which are shown in Fig. \ref{fig:Score_GDQN} and Fig. \ref{fig:Score_MVGG} respectively with average grasp success prediction during training. In Fig. \ref{fig:Score_GDQN}, all-action spaces training data have been plotted with a different colour to see the differences. The 3-action training gets faster as compared to 12 or 24 due to less no. of action predictions. The same have been observed from 12-action and 24-action; 12-action learning is faster as compared to 24-action. From Fig. \ref{fig:Score_GDQN} and Fig. \ref{fig:Score_MVGG} we infer that GDQN performs better than MVGG16. In GDQN learning initially action prediction is random but it gets stable after 400 episodes as epsilon gets decayed beyond that. The characteristics have been shown in Fig \ref{fig:epsilon} for 12-action.  and action prediction has been performed by the trained model. Fig. \ref{fig:Score_MVGG} shows that MVGG16 performance decreases drastically as its action space get increases.

\begin{figure}
\minipage{0.32\textwidth}
  \includegraphics[width=\linewidth]{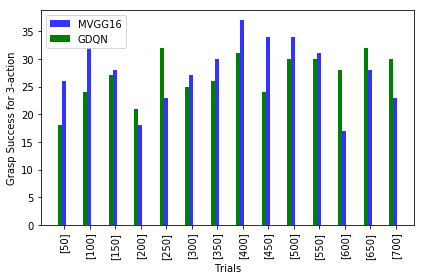}
  \caption{3-action space training}\label{fig:bar_3train}
\endminipage\hfill
\minipage{0.32\textwidth}
  \includegraphics[width=\linewidth]{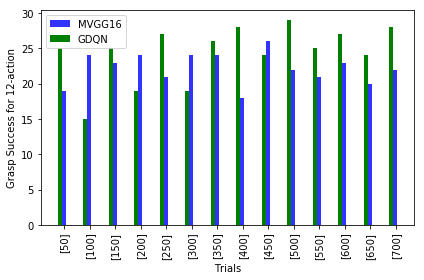}
  \caption{12-action space training}\label{fig:bar_12train}
\endminipage\hfill
\minipage{0.32\textwidth}%
  \includegraphics[width=\linewidth]{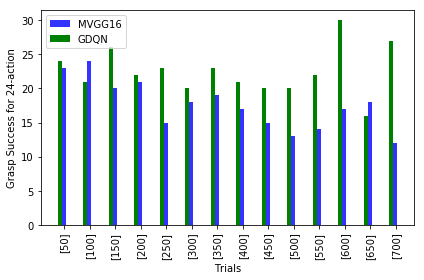}
  \caption{24-action space training}\label{fig:bar_24train}
\endminipage
\end{figure}

\begin{figure}[pos=h]
    \centering
        \includegraphics[scale=.50]{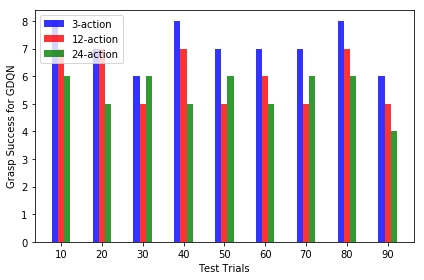}
        \caption{{Test Score by GDQN}}
    \label{fig:bar_test}
\end{figure}

Fig. \ref{fig:bar_3train},\ref{fig:bar_12train} and \ref{fig:bar_24train} have been plotted using the size of 50 batches from training episodes. Here, success rate of prediction is shown for  action space of 3, 12 and 24. It infers that as the action space increases the performance of GDQN improves.
During testing, the performance has been checked for every 10 attempts of grasping which  has been  shown in Fig. \ref{fig:bar_test}.
\\

\section{Conclusion}
In the present investigation, we have presented both machine learning and deep reinforcement learning based approaches for execution level grasping of objects having well defined shape and size.The results are quite encouraging from the point of view of easiness and simplicity, compared to Qt-opt, which is more sophisticated but requires far more computational and hardware resources.  

Through rigorous experimentation we have shown that GA based optimization predictor works well for determining the mapping matrix, which maps between vision space and robot’s configuration space.  The results show the effectiveness of GA towards generating quality solution reflected  in error convergence which converged to 0.00000207. We allowed program to predict for 10 positions as shown in the Fig. \ref{fig:GA_test} which shows for almost all the cases, the prediction was accurate. Subsequently, for orientation learning our evaluation indicates that in general GDQN outperforms MVGG16, although it is a bit slow  in learning (ref Fig \ref{fig:Score_GDQN} \& \ref{fig:Score_MVGG}), which does not matter much since the learning can be done offline. 

The robustness and stable learning for GDQN (ref Fig \ref{fig:bar_3train}, \ref{fig:bar_12train} \& \ref{fig:bar_24train}), which increases with the action space, makes it specially attractive compared to MVGG16. As mentioned earlier, in comparison to the state of the art Qt-opt for grasp solution which takes huge computational resources and hardware requirement, our method is simple, requires limited number of available data and one hardware robot only for learning and gives satisfactory results which is quite generic. Using robot for vision concept, we are also generating more and more data to be used for training to improve the future accuracy rate.

The insights suggest that the input of the center of the object to be grasped, as coming from the bounding box of the object detection algorithm, is of great significance to improve the performance of the grasp strategy developed in the robot’s configuration space. Thus a natural implication would be to explore in future the ways of getting the center point of the graspable objects more accurately by some other means such as segmentation, masking etc. 
Since the  algorithms we proposed here, all works quite well in off-policy regimes, they should reasonably work well in real time grasp manipulation as well.

\bibliographystyle{IEEEtran}
\appendix

\bibliographystyle{cas-model2-names}

\bibliography{cas-sc-template}

\begin{thebibliography}{10}
\providecommand{\url}[1]{#1}
\csname url@samestyle\endcsname
\providecommand{\newblock}{\relax}
\providecommand{\bibinfo}[2]{#2}
\providecommand{\BIBentrySTDinterwordspacing}{\spaceskip=0pt\relax}
\providecommand{\BIBentryALTinterwordstretchfactor}{4}
\providecommand{\BIBentryALTinterwordspacing}{\spaceskip=\fontdimen2\font plus
\BIBentryALTinterwordstretchfactor\fontdimen3\font minus
  \fontdimen4\font\relax}
\providecommand{\BIBforeignlanguage}[2]{{%
\expandafter\ifx\csname l@#1\endcsname\relax
\typeout{** WARNING: IEEEtran.bst: No hyphenation pattern has been}%
\typeout{** loaded for the language `#1'. Using the pattern for}%
\typeout{** the default language instead.}%
\else
\language=\csname l@#1\endcsname
\fi
#2}}
\providecommand{\BIBdecl}{\relax}
\BIBdecl

\bibitem{D1}
J.~{Bohg}, A.~{Morales}, T.~{Asfour}, and D.~{Kragic}, ``Data-driven grasp
  synthesis—a survey,'' \emph{IEEE Transactions on Robotics}, vol.~30, no.~2,
  pp. 289--309, April 2014.

\bibitem{c1}
\BIBentryALTinterwordspacing
A.~Sahbani, S.~El-Khoury, and P.~Bidaud, ``An overview of 3d object grasp
  synthesis algorithms,'' \emph{Robotics and Autonomous Systems}, vol.~60,
  no.~3, pp. 326 -- 336, 2012, autonomous Grasping. [Online]. Available:
  \url{http://www.sciencedirect.com/science/article/pii/S0921889011001485}
\BIBentrySTDinterwordspacing

\bibitem{George}
\BIBentryALTinterwordspacing
G.~Konidaris, S.~Kuindersma, R.~Grupen, and A.~Barto, ``Robot learning from
  demonstration by constructing skill trees,'' \emph{The International Journal
  of Robotics Research}, vol.~31, no.~3, pp. 360--375, 2012. [Online].
  Available: \url{https://doi.org/10.1177/0278364911428653}
\BIBentrySTDinterwordspacing

\bibitem{Peters}
L.~D. K. J. N.-T. D. B. A. S.~S. Peters, J., ``Robot learning,'' \emph{In
  Springer Handbook of Robotics, 2nd ed.; Springer: Berlin/Heidelberg, Germany,
  2017;}, pp. 357--394.

\bibitem{p27}
\BIBentryALTinterwordspacing
L.~Pinto and A.~Gupta, ``Supersizing self-supervision: Learning to grasp from
  50k tries and 700 robot hours,'' \emph{CoRR}, vol. abs/1509.06825, 2015.
  [Online]. Available: \url{http://arxiv.org/abs/1509.06825}
\BIBentrySTDinterwordspacing

\bibitem{22}
\BIBentryALTinterwordspacing
I.~Lenz, H.~Lee, and A.~Saxena, ``Deep learning for detecting robotic grasps,''
  \emph{The International Journal of Robotics Research}, vol.~34, no. 4-5, pp.
  705--724, 2015. [Online]. Available:
  \url{https://doi.org/10.1177/0278364914549607}
\BIBentrySTDinterwordspacing

\bibitem{p23}
G.~C. {Nandi}, P.~{Agarwal}, P.~{Gupta}, and A.~{Singh}, ``Deep learning based
  intelligent robot grasping strategy,'' in \emph{2018 IEEE 14th International
  Conference on Control and Automation (ICCA)}, June 2018, pp. 1064--1069.

\bibitem{6}
D.~Kragic, H.~Christensen, and F.~A, ``Survey on visual servoing for
  manipulation,'' \emph{Comput. Vis. Act. Percept. Lab. Fiskartorpsv}, vol.~15,
  02 2002.

\bibitem{D2}
\BIBentryALTinterwordspacing
S.~Levine, P.~Pastor, A.~Krizhevsky, J.~Ibarz, and D.~Quillen, ``Learning
  hand-eye coordination for robotic grasping with deep learning and large-scale
  data collection,'' \emph{The International Journal of Robotics Research},
  vol.~37, no. 4-5, pp. 421--436, 2018. [Online]. Available:
  \url{https://doi.org/10.1177/0278364917710318}
\BIBentrySTDinterwordspacing

\bibitem{1}
\BIBentryALTinterwordspacing
U.~Viereck, A.~ten Pas, K.~Saenko, and R.~P. Jr., ``Learning a visuomotor
  controller for real world robotic grasping using easily simulated depth
  images,'' \emph{CoRR}, vol. abs/1706.04652, 2017. [Online]. Available:
  \url{http://arxiv.org/abs/1706.04652}
\BIBentrySTDinterwordspacing

\bibitem{O2}
A.~{Saxena}, J.~{Driemeyer}, and A.~Y. {Ng}, ``Learning 3-d object orientation
  from images,'' in \emph{2009 IEEE International Conference on Robotics and
  Automation}, May 2009, pp. 794--800.

\bibitem{O1}
J.~Langlois, H.~Mouchère, N.~Normand, and C.~Viard-Gaudin, ``3d orientation
  estimation of industrial parts from 2d images using neural networks,'' 01
  2018, pp. 409--416.

\bibitem{Qt}
\BIBentryALTinterwordspacing
D.~Kalashnikov, A.~Irpan, P.~Pastor, J.~Ibarz, A.~Herzog, E.~Jang, D.~Quillen,
  E.~Holly, M.~Kalakrishnan, V.~Vanhoucke, and S.~Levine, ``Qt-opt: Scalable
  deep reinforcement learning for vision-based robotic manipulation,''
  \emph{CoRR}, vol. abs/1806.10293, 2018. [Online]. Available:
  \url{http://arxiv.org/abs/1806.10293}
\BIBentrySTDinterwordspacing

\bibitem{DQN}
\BIBentryALTinterwordspacing
``Deep q-networks for accelerating the training of deep neural networks,''
  \emph{CoRR}, vol. abs/1606.01467, 2016, withdrawn. [Online]. Available:
  \url{http://arxiv.org/abs/1606.01467}
\BIBentrySTDinterwordspacing

\bibitem{DQN1}
V.~Mnih, K.~Kavukcuoglu, D.~Silver, A.~Rusu, J.~Veness, M.~Bellemare,
  A.~Graves, M.~Riedmiller, A.~Fidjeland, G.~Ostrovski, S.~Petersen,
  C.~Beattie, A.~Sadik, I.~Antonoglou, H.~King, D.~Kumaran, D.~Wierstra,
  S.~Legg, and D.~Hassabis, ``Human-level control through deep reinforcement
  learning,'' \emph{Nature}, vol. 518, pp. 529--33, 02 2015.

\bibitem{G3}
D.~E. Goldberg, \emph{Genetic Algorithms in Search, Optimization and Machine
  Learning}, 2nd~ed.\hskip 1em plus 0.5em minus 0.4em\relax Addison-Wesley,
  1994.

\bibitem{OD1}
\BIBentryALTinterwordspacing
J.~Redmon and A.~Farhadi, ``Yolov3: An incremental improvement,'' \emph{CoRR},
  vol. abs/1804.02767, 2018. [Online]. Available:
  \url{http://arxiv.org/abs/1804.02767}
\BIBentrySTDinterwordspacing

\bibitem{OD2}
\BIBentryALTinterwordspacing
S.~Ren, K.~He, R.~B. Girshick, and J.~Sun, ``Faster {R-CNN:} towards real-time
  object detection with region proposal networks,'' \emph{CoRR}, vol.
  abs/1506.01497, 2015. [Online]. Available:
  \url{http://arxiv.org/abs/1506.01497}
\BIBentrySTDinterwordspacing

\bibitem{D3}
\BIBentryALTinterwordspacing
J.~Mahler, J.~Liang, S.~Niyaz, M.~Laskey, R.~Doan, X.~Liu, J.~A. Ojea, and
  K.~Goldberg, ``Dex-net 2.0: Deep learning to plan robust grasps with
  synthetic point clouds and analytic grasp metrics,'' \emph{CoRR}, vol.
  abs/1703.09312, 2017. [Online]. Available:
  \url{http://arxiv.org/abs/1703.09312}
\BIBentrySTDinterwordspacing

\bibitem{D4}
A.~Depierre, E.~Dellandr{\'e}a, and L.~Chen, ``Jacquard: A large scale dataset
  for robotic grasp detection,'' \emph{2018 IEEE/RSJ International Conference
  on Intelligent Robots and Systems (IROS)}, pp. 3511--3516, 2018.

\bibitem{p21}
P.~{Shukla} and G.~C. {Nandi}, ``Robotized grasp: Grasp manipulation using
  evolutionary computing,'' in \emph{Proceedings of 2019 International
  Conference on Electrical, Electronics and Computer Engineering (UPCON)},
  November 2019.

\bibitem{BK}
Z.~{Ju}, C.~{Yang}, and H.~{Ma}, ``Kinematics modeling and experimental
  verification of baxter robot,'' in \emph{Proceedings of the 33rd Chinese
  Control Conference}, July 2014, pp. 8518--8523.

\bibitem{atari}
\BIBentryALTinterwordspacing
V.~Mnih, K.~Kavukcuoglu, D.~Silver, A.~Graves, I.~Antonoglou, D.~Wierstra, and
  M.~A. Riedmiller, ``Playing atari with deep reinforcement learning,''
  \emph{CoRR}, vol. abs/1312.5602, 2013. [Online]. Available:
  \url{http://arxiv.org/abs/1312.5602}
\BIBentrySTDinterwordspacing

\bibitem{COCO}
\BIBentryALTinterwordspacing
T.~Lin, M.~Maire, S.~J. Belongie, L.~D. Bourdev, R.~B. Girshick, J.~Hays,
  P.~Perona, D.~Ramanan, P.~Doll{\'{a}}r, and C.~L. Zitnick, ``Microsoft
  {COCO:} common objects in context,'' \emph{CoRR}, vol. abs/1405.0312, 2014.
  [Online]. Available: \url{http://arxiv.org/abs/1405.0312}
\BIBentrySTDinterwordspacing

\bibitem{Huber}
\BIBentryALTinterwordspacing
P.~J. Huber, \emph{Robust Statistics}.\hskip 1em plus 0.5em minus 0.4em\relax
  Berlin, Heidelberg: Springer Berlin Heidelberg, 2011, pp. 1248--1251.
  [Online]. Available: \url{https://doi.org/10.1007/978-3-642-04898-2_594}
\BIBentrySTDinterwordspacing

\bibitem{Kinect}
\BIBentryALTinterwordspacing
Z.~Zhang, ``Microsoft kinect sensor and its effect,'' \emph{IEEE MultiMedia},
  vol.~19, no.~2, pp. 4--10, Apr. 2012. [Online]. Available:
  \url{https://doi.org/10.1109/MMUL.2012.24}
\BIBentrySTDinterwordspacing

\end{thebibliography}


\end{document}